\pdfoutput=1 
\documentclass{fundam}
\publyear{22}
\papernumber{2102}
\volume{185}
\issue{1}

\usepackage{url} 
\usepackage[ruled,lined]{algorithm2e}
\usepackage{graphicx}
\usepackage{tikz}
\usepackage{nomencl}
\makenomenclature
\usepackage{multicol}
\usepackage{subcaption}
\usepackage{algorithm2e}
\usepackage{amsmath}
\usepackage{hyperref}
\usepackage{epstopdf}

\usetikzlibrary{automata, positioning, arrows}

\begin{document}

\title{Log Optimization Simplification Method for Predicting Remaining Time}

\author{
    Jianhong Ye\thanks{Address for correspondence: College of Computer Science and Technology, Huaqiao University, Jimei Avenue 668, 361021 Xia’men, Fujian, China.
}, Siyuan Zhang\\
    College of Computer Science and Technology, Huaqiao University\\
    361021 Xia’men, Fujian, China\\
    leafever{@}hqu.edu.cn\\
    24014083067{@}stu.hqu.edu.cn
    \and Yan Lin
    \\Jiangsu Branch, Agricultural Bank of China
    \\ 210000 Nan’jing, Jiangsu, China
} 

\maketitle
\runninghead{J. Ye et al.}{Log-Based Time Prediction Method}

\begin{abstract}
Information systems generate a large volume of event log data during business operations, much of which consists of low-value and redundant information. When performance predictions are made directly from these logs, the accuracy of the predictions can be compromised. Researchers have explored methods to simplify and compress these data while preserving their valuable components. Most existing approaches focus on reducing the dimensionality of the data by eliminating redundant and irrelevant features. However, there has been limited investigation into the efficiency of execution both before and after event log simplification. In this paper, we present a prediction point selection algorithm designed to avoid the simplification of all points that function similarly. We select sequences or self-loop structures to form a simplifiable segment, and we optimize the deviation between the actual simplifiable value and the original data prediction value to prevent over-simplification. Experiments indicate that the simplified event log retains its predictive performance and, in some cases, enhances its predictive accuracy compared to the original event log.
\end{abstract}

\begin{multicols}{2}
\printnomenclature
\end{multicols}
\nomenclature[01]{$A$}{the set of activities}
\nomenclature[02]{$C^{'}$}{the set of nodes}
\nomenclature[03]{$e$}{event}
\nomenclature[04]{$E$}{the set of (possible) events}
\nomenclature[05]{$Elog_i$}{the event log of }
\nomenclature[06]{$F$}{the set of arcs}
\nomenclature[07]{$g$}{a slack variable}
\nomenclature[08]{$GSPN$}{generalised stochastic Petri net}
\nomenclature[09]{$L$}{event log}
\nomenclature[10]{$MAE$}{the error in predicting the remaining time}
\nomenclature[11]{$N^*$}{the set of reducible substructures}
\nomenclature[12]{$NElog$}{a log generated by the system after the structure simplification}
\nomenclature[13]{$P$}{the set of places}
\nomenclature[14]{$P_{r}$}{the set of prediction points}
\nomenclature[15]{$PN$}{Petri net}
\nomenclature[16]{$R^{'}$}{the set of relations}
\nomenclature[17]{$S$}{resource community network}
\nomenclature[18]{$t_{\gamma}^{\sigma}$}{the execution time of the activity  $\gamma$  in $\sigma$}
\nomenclature[19]{$T$}{the set of transitions}
\nomenclature[20]{$W,W^{'}$}{the weight function}
\nomenclature[21]{$\sigma$}{trace}
\nomenclature[22]{$\gamma$}{activity}
\nomenclature[23]{$T$}{the set of all possible traces}
\nomenclature[24]{$\Gamma$}{the expected deviation between the predicted value and the true value}
\nomenclature[25]{$\mu_i$}{the deviation value}
\nomenclature[26]{$Q$}{modularity}
\nomenclature[27]{$\Delta Q$}{modularity gain}

\section{Introduction}
\begin{figure}[htbp]
    \centering
    \begin{subfigure}{\linewidth}
    \includegraphics[width=\linewidth]{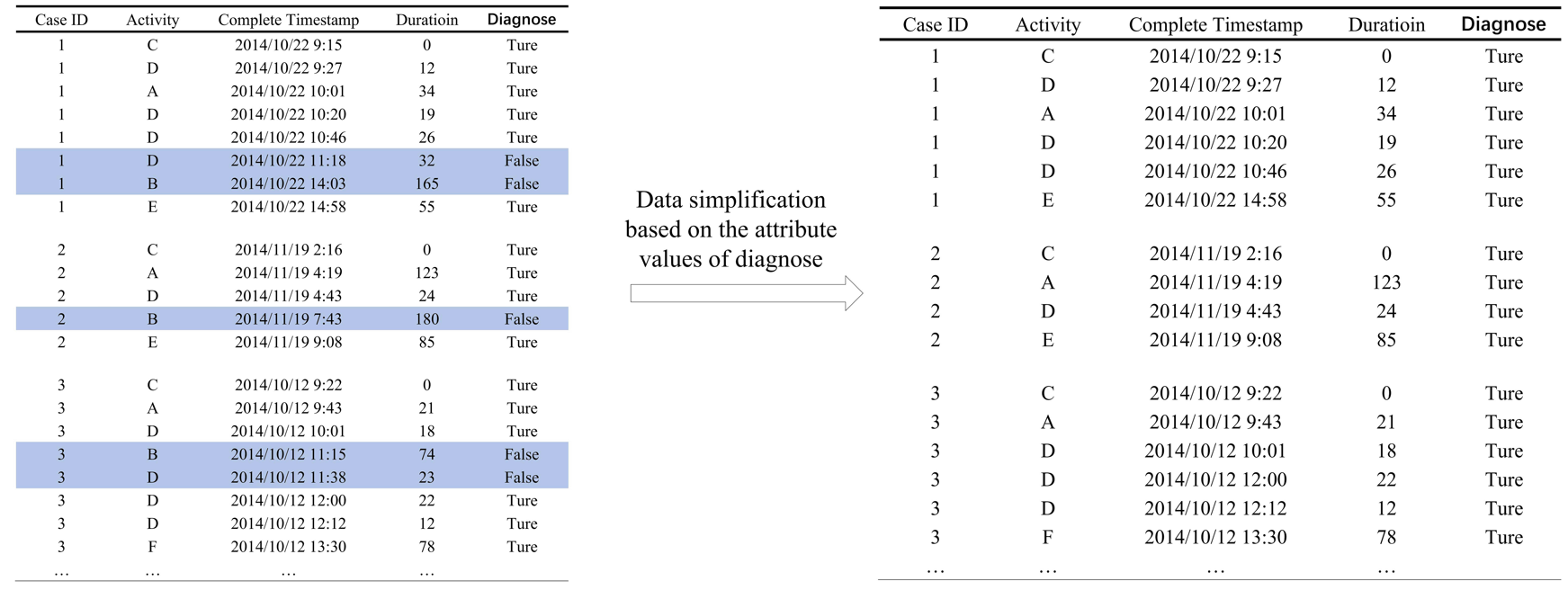} 
    \caption{Simplification based on the attribute value of diagnose}
    \label{fig:figure 1(a)}
    \end{subfigure}
    \begin{subfigure}{\linewidth}
    \includegraphics[width=\linewidth]{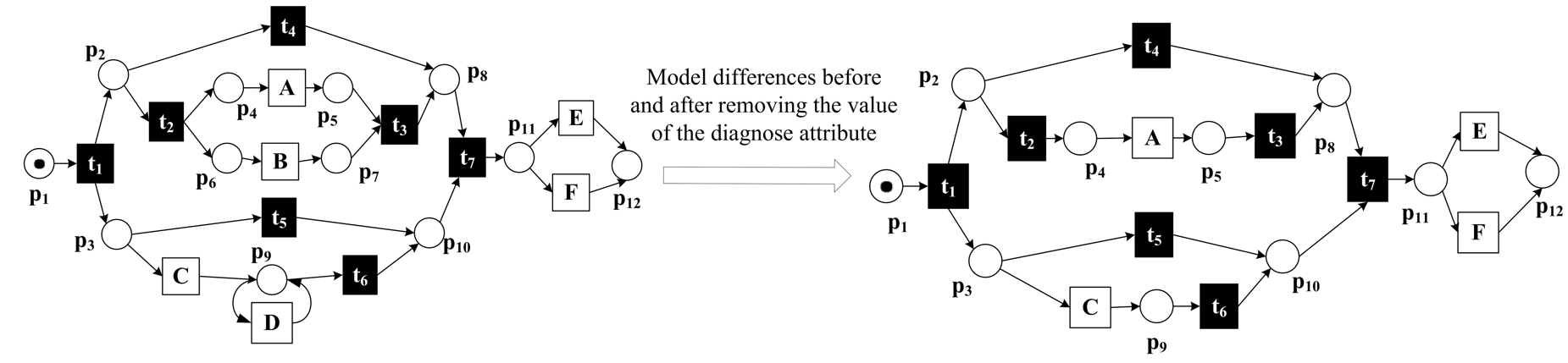} 
    \caption{Remove before and after model differences where the diagnosis is false}
    \label{fig:figure 1(b)}
    \end{subfigure}
    \caption{Modified the process model to simplify logs by focusing on specific attributes.}
    \label{fig:figure 1}
\end{figure}
Predictive monitoring involves taking appropriate actions based on the predictions provided by the system. This can include rationally planning existing resources or adjusting the priority of services according to the current operational status. The goals of prediction may include estimating the remaining time (Syamsiyah, van Dongen, and van der Aalst 2017 \cite{syamsiyah2017discovering}; Ladleif and Weske 2020 
\cite{ladleif2020time}; Cesario et al. 2016 \cite{cesario2016cloud}; Verenich et al. 2019 \cite{verenich2019survey}; Verenich et al. 2019 \cite{verenich2019survey}; Cao et al. 2023 \cite{cao2023explainable}; Elyasi, van der Aalst, and Stuckenschmidt 2024 \cite{elyasi2024pgtnet}), assessing risks (De Leoni, Dees and Reulink 2020\cite{de2020design}; Pika et al. 2016 \cite{pika2016evaluating}; Metzger and Bohn 2017\cite{metzger2017risk}; Teinemaa et al. 2018 \cite{teinemaa2018alarm}; Cardoso, Respício, and Domingos 2024 \cite{cardoso2024granular}), predicting the next activity (Tax et al. 2017 \cite{tax2017predictive}; ; Taymouri et al. 2020 \cite{taymouri2020predictive}; Kaftantzis et al. 2024 \cite{kaftantzis2024predictive}; Sun et al. 2024\cite{sun2024next}), or forecasting specific indicators (either single or aggregated) (Folino, Folino, and Pontieri 2018 \cite{folino2018ensemble}; Di Francescomarino et al. 2018 \cite{di2018predictive}; Teinemaa et al. 2019 \cite{teinemaa2019outcome}), among others. Increasingly, systems are looking to process mining as a way to support online modeling and analysis, enabling rapid establishment of predictive process monitoring and simplifying logs as an efficient method. There are various dimensionality reduction techniques available, such as principal component analysis, singular value decomposition, latent semantic analysis, linear discriminant analysis, multidimensional scaling, and learning vector quantization. These methods can be applied across different fields. 
Log simplification can also be approached from a model perspective (Tsagkani and Tsalgatidou 2022 \cite{tsagkani2022process}). Tax et al. describe a method for abstracting low-level events in event logs using supervised learning. In this method, supervised event-abstracted synthetic logs are employed to discover smaller, more comprehensible high-level models (Tax et al. 2017 \cite{tax2017predictive}). However, their approach struggles when dealing with logs that contain numerous repeated events. This is because the long short-term memory (LSTM) predictions can be excessively prolonged, resulting in an overestimation of remaining cycle times. Teinemaa et al. proposed a method for configuring process abstraction with a specific abstraction goal (Teinemaa et al. 2019 \cite{teinemaa2019outcome}). A significant limitation of this method is its lack of interpretability regarding predictions. Fahland et al. applied process post-processing techniques to simplify the discovered process models. Their approach relies on branching processes to address the problems of overfitting and underfitting (Fahland and van der Aalst 2013 \cite{fahland2013simplifying}). Upreti utilized dimensional analysis and model fitting methods to approximate the models (Upreti 2017 \cite{upreti2017process}). Meanwhile, Senderovich et al. conducted a series of folding operations to simplify the model's structure and improve prediction accuracy ( Senderovich et al. 2018 \cite{senderovich2018aggregate}). However, they did not consider how changes in the event log might affect their outcomes. Figure \ref{fig:figure 1} illustrates how data is deleted based on the attribute value of ''diagnose''. It shows that activities B and D are removed from the model due to the loss of the corresponding attribute.

\begin{figure*}[htbp]
    \centering
    \includegraphics[width=0.6\textwidth]{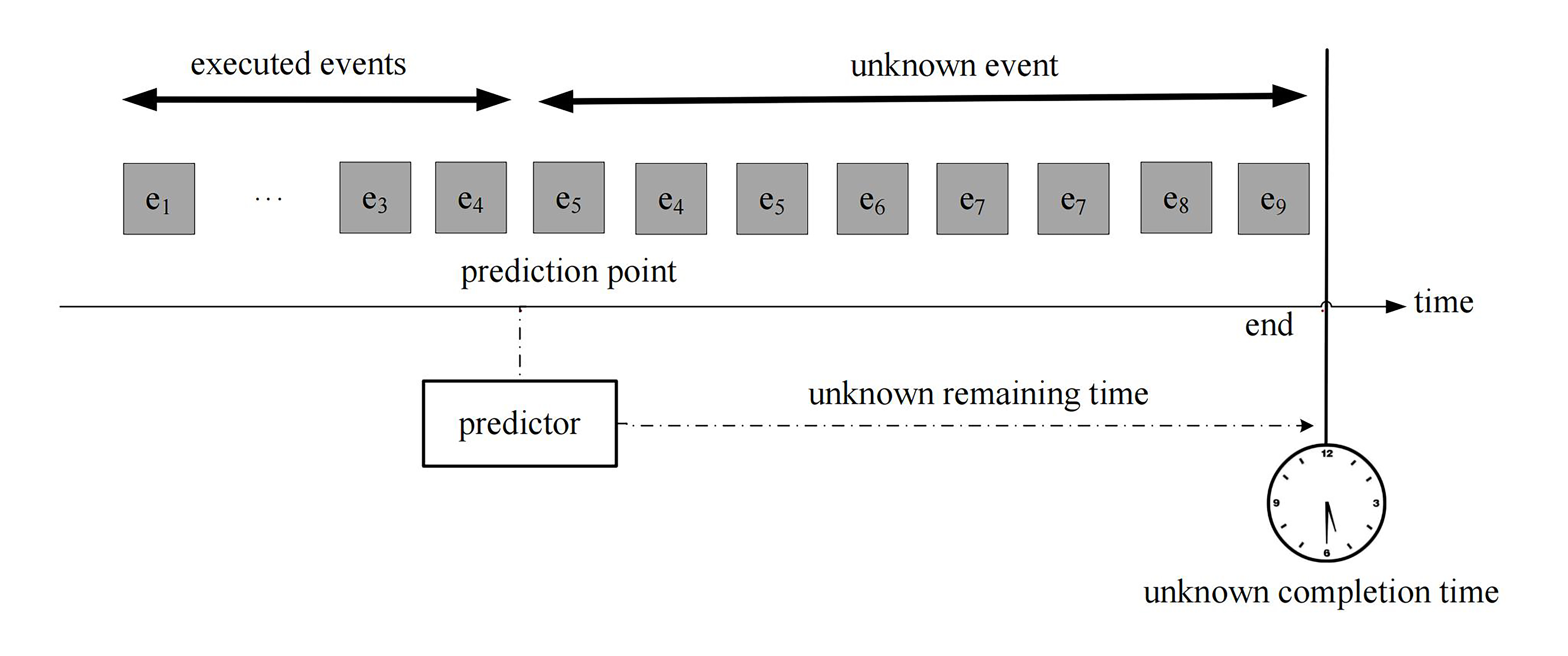} 
    \caption{Estimate remaining time.}
    \label{fig:figure 2}
\end{figure*}
Time is a crucial factor in business processes. If it is possible to predict the remaining time of an important event in the current situation, arrangements can be made in advance to enhance the company's work efficiency. Figure \ref{fig:figure 2} illustrates the time remaining from the prediction point to the event's end. Events prior to the prediction point have already been executed, while events following the prediction point remain uncertain. To achieve the prediction objective, this paper employs a log simplification based on the process model. The approach preserves as much valuable information as possible and emphasizes the prediction of remaining time.

The innovation of this paper lies in its comprehensive consideration of how simplification impacts prediction performance. Traditional simplification methods typically focus solely on the effects of attribute and structure deletion on prediction results. This paper presents two key innovations. First, it introduces simplification constraints for prediction points. Certain important prediction points cannot be deleted, as their removal would significantly impair prediction performance. To address this, we utilize resource community networks to cluster nodes that share similar roles. Nodes within the same community are selected as prediction points and cannot be removed simultaneously from the substructure. This aspect is discussed in Chapter \ref{sec3}. The second innovation is the optimization analysis conducted prior to simplifying the substructure. If it is determined that simplifying the substructure would detrimentally affect prediction performance, adjustment operations are made. This aspect is detailed in Chapter \ref{sec4}.

\section{Definition}\label{sec2}

Let $e$ be the event that occurs during the execution of a process. Let $T$ be the set of all possible traces defined as a sequence of events, such that, $\sigma \in T, \sigma = <e_1,e_2,\ldots,e_n> $, an event log can be defined as a set of traces, $L \subseteq T$ , where $L=\{ \sigma_1,\sigma_2,\ldots,\sigma_n \}$. Let $A$ be a set of activities(i.e. tasks in this paper)corresponding to the set of transitions and a set of agents (i.e. resources, individuals or workers). $E = A\times P$ be the set of (possible) events(i.e. combinations of an activity and an agent). Table \ref{tab:table 1} shows an example of such as an event log.
The tuple $PN=(P,T,F,W,A)$ represents a generalised stochastic Petri net (\textbf{GSPN}):
\begin{itemize}
    \item $P$ :represents the set of places of Petri nets.
    \item $T$ :$T_t \cup T_{\varepsilon} $ , represents the set of transitions of Petri nets; $T_t$,  represents the set of visible transitions; $T_{\varepsilon}$, represents the set of invisible transitions.
    \item $F \subseteq (P \times T) \cup (T \times P)$ : represents the set of arcs with directional Petri net variation.
    \item $W$ : $F \to N$, represents the weight function of the directed Petri net arcs, $N=(1,2,3,\ldots)$ .
    \item $A$ :represents the set of activities.
\end{itemize}

\textbf{Modularity} is a metric used to evaluate the quality of clusters within a network. It is defined as follows:

\begin{equation}
    \label{sec_equ1}
    Q = \frac{1}{{2m}} \sum\limits_{i,j} \left( W(p_i, p_j) - \frac{k_i k_j}{2m} \right) \cdot \delta(C_x, C_y)
\end{equation}

In this equation, \(0 < i, j < |P|\) and \(i \neq j\), while \(0 < x, y < |C'|\). Here, \(W(p_i, p_j)\) represents the weight of the edge connecting the nodes \(p_i\) and \(p_j\), where \(p_i, p_j \in P\). The term \(k_i = \sum\limits_{1 \leq j \leq |P|} W(p_i, p_j)\) denotes the total weight of all edges connected to node \(p_i\). 

The function \(\delta(C_x, C_y)\) equals 1 if nodes \(p_i\) and \(p_j\) are clustered within the same community, \(C_x\) and \(C_y\), respectively; otherwise, it equals 0. Finally, \(m = \sum\limits_{1 \leq i, j \leq |P|} W(p_i, p_j)\) is the total weight of all edges in the social network. In general, the value of \(Q\) ranges from 0 to 1.

When a node \( p_i \) in community \( C_x \) is transferred to another community \( C_y \), the change in modularity is known as the \textbf{modularity gain} and is denoted as \( \Delta Q^i_{xy} \). In this notation, the superscript ``$i$'' specifies the particular node \( p_i \), while the subscript ``$xy$'' indicates the direction of the movement from community \( C_x \) to community \( C_y \). The modularity gain from this transfer is calculated using Eq. (\ref{sec_equ2}):
\begin{figure}
	\begin{equation}
		\begin{aligned}
			\label{sec_equ2}  \Delta Q^i_{xy} = \left( {\frac{\sum
					\limits_{1\leq
						k,l \leq |C_y|} W(p_k, p_l)+ k_i^y}{2m}} -\left( \frac{\sum \limits_{\substack{1\leq k \leq |C_y| \\
						1\leq j \leq |P-C_y|}}W(p_k, p_j) + k_i}{2m}\right)^2 \right) - \\
			\left(\frac{\sum \limits_{1\leq k,l \leq |C_y|} W(p_k, p_l)}{2m} -
			\left( \frac{\sum \limits_{\substack{1\leq k \leq |C_y| \\
						1\leq j \leq |P-C_y|}}W(p_k, p_j)}{2m} \right)^2- \left(
			\frac{k_i}{2m}\right)^2 \right)
		\end{aligned}
	\end{equation}
\end{figure}

Let \( C_y \) represent a community in a social network \( G \). The term \( \sum \limits_{1 \leq k, l \leq |C_y|} W(p_k, p_l) \) denotes the total weight of the edges within the community \( C_y \). Furthermore, \( \sum \limits_{\substack{1 \leq k \leq |C_y| \\ 1 \leq j \leq |P - C_y|}} W(p_k, p_j) \) represents the sum of the weights of the edges connecting the nodes in the community \( C_y \) to the nodes in other communities. The variable \( k_i \) is defined as \( k_i = \sum \limits_{1 \leq l \leq |P|} W(p_i, p_l) \), indicating the total weight of all edges connected to node \( p_i \). 

Moreover, we can derive \( k_i^y = \sum \limits_{1 \leq k \leq |C_y|} W(p_i, p_k) \), which restricts the sum of edge weights in the community \( C_y \) that connect to the node \( p_i \). Similarly to the description of modularity, \( m \) is the total weight of all edges in the social network \( G \).

Eq. (\ref{sec_equ2}) describes the difference in modularity for the community \( C_y \) when node \( p_i \) is included versus when it is excluded. This equation can be simplified to 

\[
\Delta Q^i_{xy} = \frac{k_i^y}{2m} - \frac{\left( \sum \limits_{\substack{1 \leq k \leq |C_y| \\ 1 \leq j \leq |P - C_y|}} W(p_k, p_j) \right) \cdot k_i}{2m^2}.
\] 

This formulation clearly outlines the relationships between the internal and external connections of the community and their impact on modularity.

We define an undirected graph $S=(C^{'},R^{'},W^{'})$ as a \textbf{resource community network}. Here, $C^{'}$ is the set of nodes, $R^{'} \subseteq C^{'} \times C^{'}$ is the set of relationships, and $W^{'}$ represents the weight function. Each node in $C^{'}$ is constructed from multisets of actors over a set \( P \). The weight function $W^{'}$ is defined such that the weight between different communities is equal to the sum of the weights of the existing edges between those communities. Conversely, the weight of an individual community is determined by the sum of the weights of the existing edges within that community. 
If there exist \( p_i \in C_i \) and \( p_j \in C_j \) where \( C_i, C_j \in C' \), then the weight function is defined as:
$W^{'}(C_i,C_j)=\sum_{(p_{i},p_{j} \in R)} W(P_i,P_j)$.

Let \( L \) be an event log, where $t_{\gamma}^{\sigma}$ represents the \textbf{execution time} of activity \( \gamma \) within the context of case \( \sigma \), encompassing both working time and waiting time. If \( L \) contains cases \( \{\sigma_1, \sigma_2, \ldots, \sigma_n\} \) where each \( \sigma_i \) belongs to \( L \), then \( Elog_i \) denotes the event log related to \( \sigma_i \), and \( e_i \) is an activity within \( \sigma_i \). When making predictions about system performance at \( e_i \), another activity \( e_j \) is referred to as a \textbf{prediction point}. The network structure reconstructed from the log \( Elog_i \) is denoted as \( N_i \). Following the simplification of the structure \( N_i \), a new log is generated, referred to as \( NElog_i \). The \textbf{deviation value} indicates the predicted power deviation between the two prediction points. For instance, if \( e_i \) and \( \hat{e}_i \) are the prediction points for \( Elog_i \) and \( NElog_i \) respectively, the deviation value can be calculated as \( \mu_i = e_i - \hat{e}_i \).
\begin{table}
    \centering
    \caption{An event log}
    \begin{tabular}{c|c|c}
    \textbf{Case identifier} & \textbf{Activity identifier} & \textbf{Resource}\\
    \hline
    Case1 & Activity A & John\\ 
    Case2 & Activity A & John\\ 
    Case3 & Activity A & Sue\\ 
    Case3 & Activity B & Carol\\ 
    Case1 & Activity B & Mike\\ 
    Case1 & Activity C & John\\ 
    Case2 & Activity C & Mike \\ 
    Case4 & Activity A & Sue \\ 
    Case2 & Activity B & John\\ 
    Case2 & Activity D & Pete\\ 
    Case5 & Activity A & Sue\\ 
    Case4 & Activity C & Carol\\ 
    Case1 & Activity D & Pete\\
    Case3 & Activity C & Sue\\
    Case3 & Activity D & Pete\\
    Case4 & Activity B & Sue\\
    Case5 & Activity E & Clare\\
    Case5 & Activity D & Clare\\
    Case4 & Activity D & Pete\\
    \end{tabular}
    \label{tab:table 1}
\end{table}
\section{Prediction point based on Resource Community Network}\label{sec3}
\subsection{Resource Community Network}

\begin{figure*}[htbp]
	\centering
	\begin{subfigure}{0.45\linewidth}
		\centering
		\includegraphics[width=\linewidth]{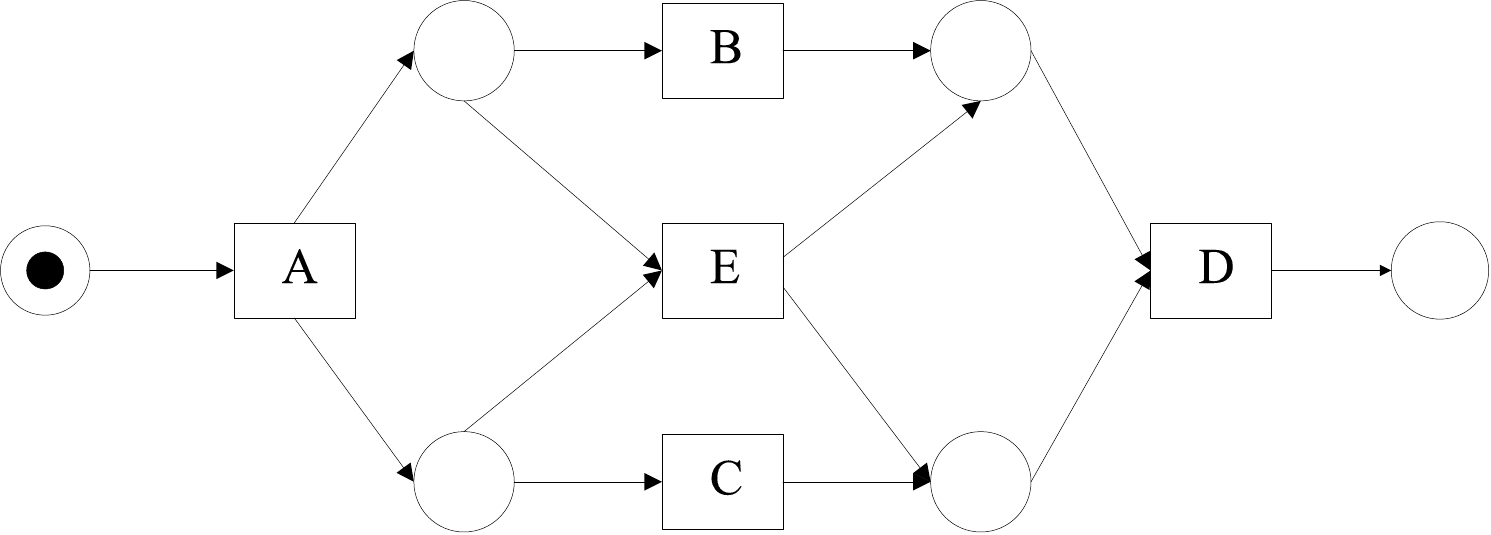}
		\caption{Petri Net}
            \label{fig:figure 4(a)}
	\end{subfigure}
	\begin{subfigure}{0.45\linewidth}
		\centering
		\includegraphics[width=0.8\linewidth]{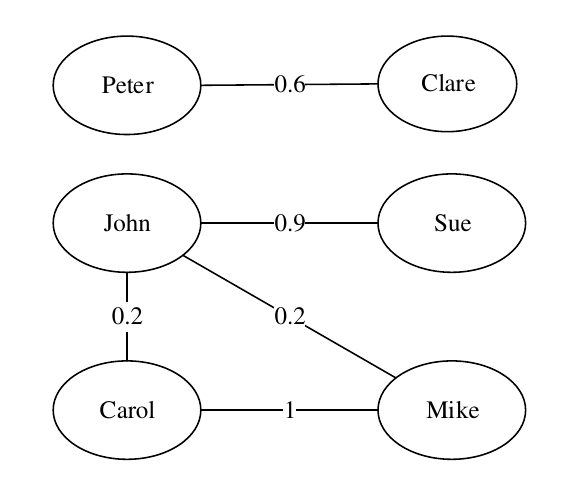}
		\caption{Social network}
            \label{fig:figure 4(b)}
	\end{subfigure}
        \caption{A Petri net and social network generated from Table \ref{tab:table 1}}
	\label{fig:figure 4}
\end{figure*}

Figure \ref{fig:figure 4(a)} shows the Petri net model generated by applying the $\alpha$-algorithm to the data in Table \ref{tab:table 1}. Additionally, Figure \ref{fig:figure 4(b)} illustrates the connection between the performers and the activities, as represented in the resulting social network.

A social network reflects the strength of correlations among its individuals. Typically, certain performers, known as resource nodes, fulfill similar roles within the network, allowing them to form their own communities, referred to as resource community networks. To achieve this, we will utilize modularity gain. Our resource community network emphasizes the concept of groups, focusing on how to create a community of individuals with similar roles. Using a social network as a foundation, Algorithm \ref{Alg1} initiates the formation of communities, with each node belonging to a single community at the outset.

For a node \( p_j \) in the social network, we consider which neighbor would be the best fit for it. 
Without loss of generality, we assume that a node \( p_i \) in community \( C_x \) is attempting to become the neighbor of node \( p_j \). A modularity gain, denoted as \( \Delta Q^i_{xy} \), is calculated according to Eq. (\ref{sec_equ2}). If there are multiple nodes attempting to become neighbors of \( p_j \), and among these nodes, \( \Delta Q^i_{xy} \) has the maximum positive value, then node \( p_i \) will be transferred to community \( C_y \). Otherwise, \( p_i \) will remain in its current community. In this recursive process, communities with several clustered nodes can be treated as new entities. Each new entity will have a loop edge, where the weight of the loop edge is the sum of the weights of the internal nodes. Additionally, the weight between different communities is determined by the sum of the weights of the nodes connecting them. Algorithm \ref{Alg1} will terminate once no nodes transfer between communities. The result is the formation of a resource community network. The time complexity of Algorithm \ref{Alg1} is \( O(n \times (n + n^2)) \), which simplifies to \( O(n^3) \).

By applying Algorithm \ref{Alg1} to the social network depicted in Figure \ref{fig:figure 4(b)}, we consider the order of traversing the nodes as follows: \emph{John}, \emph{Sue}, \emph{Mike}, \emph{Carol}, \emph{Peter}, and \emph{Clare}. First, let us examine the possibility of \emph{John} joining other communities. \emph{John} has three neighbors: \emph{Sue}, \emph{Mike}, and \emph{Carol}. If \emph{John} decides to join \emph{Sue}'s community, the modularity gain can be calculated using the formula:
\[
\Delta Q^i_{xy} = \frac{k_i^y}{2m} - \frac{\left(\sum \limits_{\substack{1 \leq k \leq |C_y| \\ 1 \leq j \leq |P-C_y|}} W(p_k, p_j)\right) \cdot k_i}{2m^2}
\]
\begin{algorithm}[htb]{\tiny}
\caption{ Discovery A Resource community network} 
\label{Alg1}
\begin{small}
    \KwIn {An initial social network \emph{G}=(\emph{P},\emph{R},\emph{W}).}
    \KwOut { A resource community network ${S}$=$({C}^{\prime},{R}^{\prime},{W}^{\prime})$.}
    \While {nodes are moving between communities}{
    $C_y \gets \{p_j|p_j \in P\}$\;
    \For{$P_k \in P$}{
        \If {$(p_{k},p_{j}) \text{ or } (p_j,p_k) \text{ in } R$}
        {calculate $\Delta Q_{xy}^{k}$ according to Eq.\ref{sec_equ2}\;}
    }
    Choose $ max \Delta Q_{xy}^{k}$\;
    $C_y \gets p_i$\; 
    ${G}$ is updated to a new social network $G^{\prime}$, and $\{p_i,p_j\}$ is considered a node in $G^{\prime}$\;
    The set \(\{p_i, p_j\}\) contains a loop edge with a weight of \(W(p_i, p_j)\)\;
    The weights between $\{p_i, p_j\}$ and other nodes are determined by the sum of the weights of the connections to $p_j$ and $p_i$.\;
    }
\end{small}
\end{algorithm}

According to Eq. (\ref{sec_equ2}), where \( m \) is the sum of the weights of the edges in the social network \( G \), given by \( m = \sum\limits_{1 \leq i,j \leq |P|} W(p_i,p_j) = 0.6 + 0.2 + 0.9 + 0.2 + 1 = 2.9 \). Here, \( k_i^y \) represents the sum of the weights of edges in \emph{Sue}'s community that are connected to node \( p_i \) (i.e., \emph{John}). Therefore, we have \( k_i^y = 0.9 \). Additionally, \( k_i \) denotes the sum of the weights of all edges in \( G \) connected to node \emph{John}, which calculates to \( k_i = 0.9 + 0.2 + 0.2 = 1.3 \). The term \( \sum \limits_{\substack{1 \leq k \leq |C_y| \\ 1 \leq j \leq |P-C_y|}} W(p_k, p_j) \) represents the sum of the weights of the edges connected to the nodes in \emph{Sue}'s community, yielding a result of 0.9. Consequently, the modularity gain for \emph{John} moving to \emph{Sue}'s community is:
\[
\Delta Q_{John \rightarrow Sue's} = \frac{0.9}{2 \times 2.9} - \frac{0.9 \times 1.3}{2 \times 2.9^2} = 0.085
\]
Next, if \emph{John} wishes to join \emph{Mike}'s community, the modularity gain can be computed as:
\[
\Delta Q_{John \rightarrow Mike's} = \frac{0.2}{2 \times 2.9} - \frac{1.2 \times 1.3}{2 \times 2.9^2} = -0.058
\]
For \emph{John} joining \emph{Carol}'s community, the modularity gain is also:
\[
\Delta Q_{John \rightarrow Carol's} = -0.058
\]

\begin{figure}[htbp]
    \centering
    \includegraphics[scale=0.8]{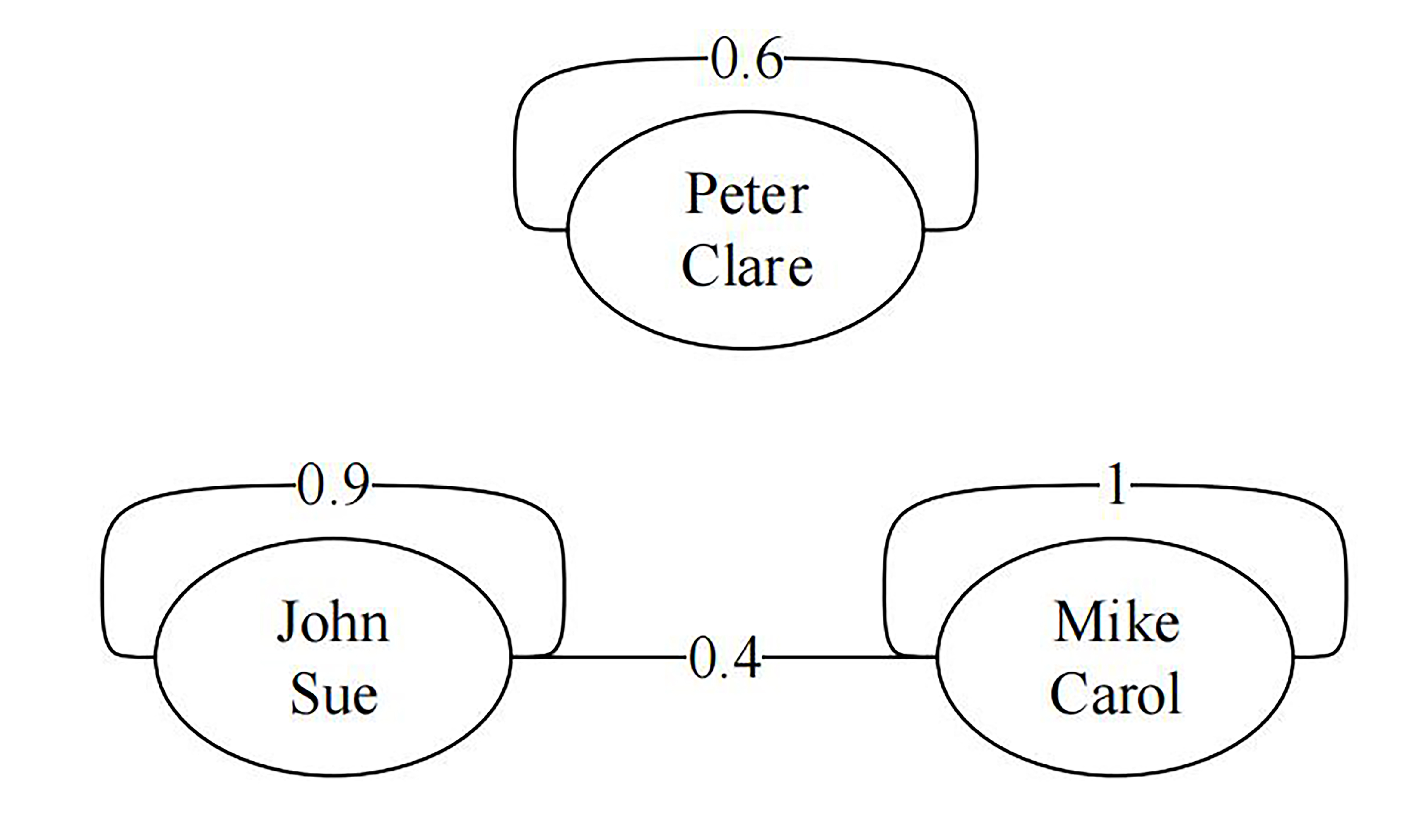} 
    \caption{A resource community network developed based on Figure \ref{fig:figure 4} and Table \ref{tab:table 1}}
    \label{fig:figure 5}
\end{figure}

Since \( 0.085 > -0.058 \), \emph{John} can indeed join \emph{Sue}'s community to form a new community. The following steps are similar, leading us to the resource community network illustrated in Figure \ref{fig:figure 5}. From Figure \ref{fig:figure 4(b)} to Figure \ref{fig:figure 5}, the key difference is our focus on groups rather than individuals. By applying the principle of maximizing modularity gain, we successfully construct three communities:\( \{Peter, Clare\} \), \( \{John, Sue\} \), and \( \{Carol, Mike\} \).

\subsection{Selection of Prediction Points}
A resource community network reflects the commonalities among individuals' resources. Nodes belonging to the same resource community network should not be deleted simultaneously during future simplifications, as all nodes within a community share similar roles. It is crucial to ensure that at least one node in each resource community network is designated as an undeletable prediction point. This process is illustrated in Algorithm \ref{Alg2}.

\begin{algorithm}[htb]{\tiny}
\caption{Selecting Prediction points}
\label{Alg2}
    \KwIn {A resource community network $S=(C^{\prime},R^{\prime},W^{\prime})$.}
    \KwOut {A set of prediction points $Pr$}
    Initialize each resource community $C_{i},C_i\subseteq C^{\prime},i=1,2,\ldots,|C^{\prime}|$\;
    \For{ $j \leftarrow 1$ to $|C^{\prime}|$}{
        \For {$p_k \in C_j$ }{
            Determine which activity node $p_k$ belongs to within the activity set $A$\;
        }
        Generate the corresponding set of node activities, denoted as \( A_j \)\;
    }
    \For{ $j \leftarrow 1$ to $|C^{\prime}|$}{Collect nodes from each $A_j$ to create a unique $Pr$ set, ensuring that no node is repeated and each comes from a different $A_j$\;}
\end{algorithm}

According to Figure \ref{fig:figure 5}, there are three communities \emph{\{John, Sue\}}, \emph{\{Peter, Clare\}} and \emph{\{Carol, Mike\}}. The activity node set corresponding to \emph{John} and \emph{Sue} is \emph{\{A, B, C\}}. Similarly, the activity node sets corresponding to resource communities \empty{\{Peter, Clare\}} and \emph{\{Carol, Mike\}} are \emph{\{D, E\}} and \emph{\{B, C\}} respectively. We need to select at least three prediction points from these three sets of activity nodes. For instance, if we choose node \emph{A} from the set \emph{\{A, B, C\}}, we can then select one node from the other sets: \empty{D} from \emph{\{D, E\}} and \emph{C} from \emph{\{B, C\}}. Applying Algorithm \ref{Alg2}, we obtain three prediction points:\emph{A, D}, and \emph{C}, each belonging to different resource communities. Prediction points should typically not be removed during the simplification process.

\section{Reduction Based on Estimated Remaining Time}  \label{sec4}
\subsection{Substructure Simplification}
\begin{figure*}[htbp]
    \centering
    \begin{minipage}{0.5\linewidth}
        \centering
        \begin{subfigure}[b]{\linewidth}
            \centering
            \includegraphics[width=\linewidth]{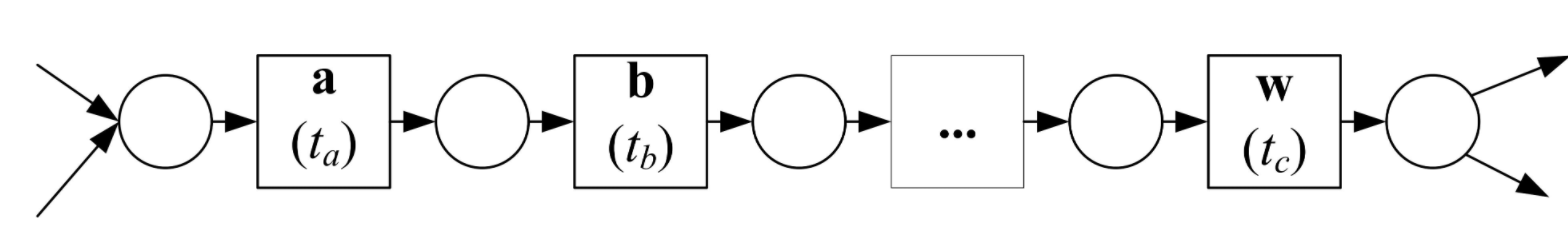}
            \caption{\emph{Sequence} structure}
            \label{fig:figure 6(a)}
        \end{subfigure}
        \vspace{0.5em}
        \begin{subfigure}[b]{\linewidth}
            \centering
            \includegraphics[width=\linewidth]{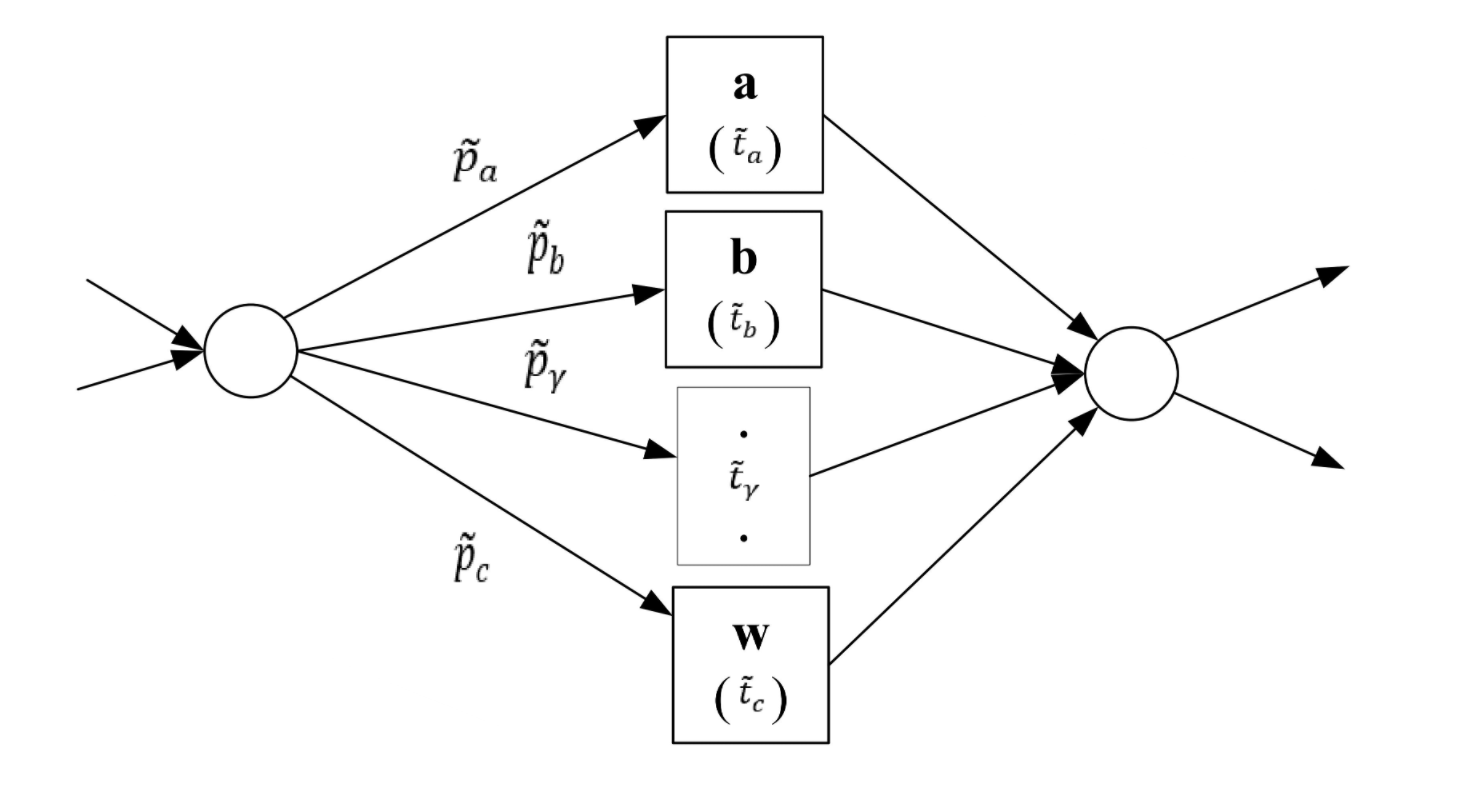}
            \caption{\emph{Or} structure}
            \label{fig:figure 6(b)}
        \end{subfigure}
        \vspace{0.5em}
        \begin{subfigure}[b]{\linewidth}
            \centering
            \includegraphics[width=\linewidth]{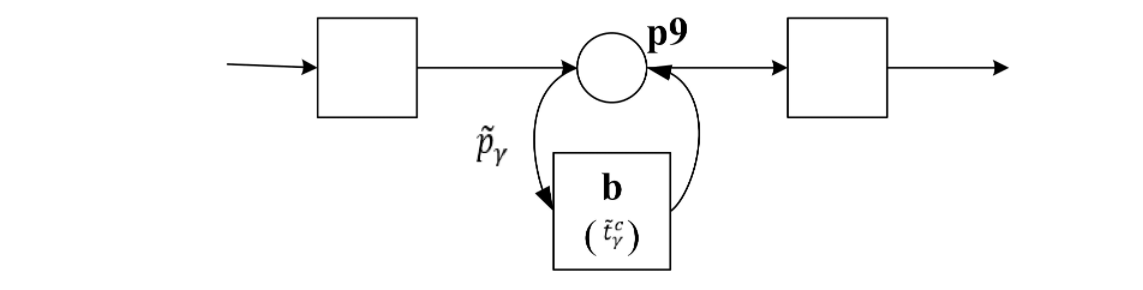}
            \caption{\emph{Self-loop} structure}
            \label{fig:figure 6(c)}
        \end{subfigure}
    \end{minipage}
    \begin{minipage}{0.4\linewidth}
        \centering
        \includegraphics[width=0.35\linewidth]{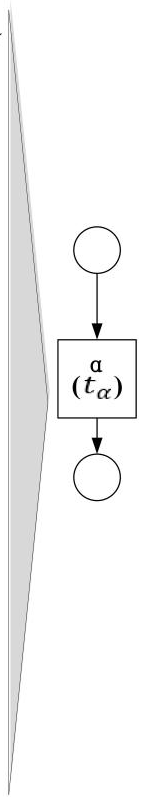}
        \label{fig:right_side_structure}
    \end{minipage}
    \caption{Simplify the structures associated with each substructure.}
    \label{fig:figure 6}
\end{figure*}

In Chapter \ref{sec3}, the selection of prediction points imposes constraints on which nodes can be simplified in Chapter \ref{sec4}. This section must consider the actual simplification process. Since both the wait time and service time of the Generalized Stochastic Petri Net (GSPN) stay within the transition, this paper focuses solely on subnet identification and the simplification of the transition. In addition to structural folding, this simplification also takes into account its impact on prediction performance. As illustrated in Figure \ref{fig:figure 6}, three types of subnet structures are identified:\emph{Sequence, Or, } and \emph{Self-loop structure}.

\begin{enumerate}
\item Simplification of the \emph{sequence} subnet \\
As shown in Figure \ref{fig:figure 6(a)}, the sequence  $\sigma_i \text{ is} <a,b,\ldots,w>$ and has only one input and one output. The execution time of the new activity $\alpha$ that replaces the sequence $\sigma_i$ after the simplification is:
\begin{equation}\label{sec_equ3}
t_{\alpha}^{\sigma_i} = \sum_{\gamma=\alpha}^{w}t_{\gamma}^{\sigma_i}
\end{equation}
    \item Simplification of the \emph{or} subnet \\
 As shown in Figure \ref{fig:figure 6(b)}, the or subnet contains activities $a$ through $w$. Each time the or structure is executed, a transition is randomly selected to occur, so the sequence of transitions involved is a set $L=\{\sigma_1,\sigma_2,\ldots,\sigma_n\}$. All logs associated with that set or structure are replaced by a new transition $t_{\alpha}^{L}$. The time spent on the or structure in all previous logs is:
\begin{equation}\label{sec_equ_4}
t_{\alpha}^{L} = \frac{\sum_{i=1}^{n} t_{\gamma,\gamma \in \{a,b,\ldots,w\}}^{\sigma_i}}{n}
\end{equation}
    \item Simplification of the \emph{self-loop} subnet \\
 In Figure \ref{fig:figure 6(c)}, assuming a self-loop occurs at transition $a$,the sequence $\sigma_i \text{is} <a,bm,\ldots,w>$,and transition $a$ occurs $m$ times. In the simplified protocol, $a$ is replaced by a new transition $t_{\alpha}^{\sigma_i}$, and the delay of the new transition is:
\begin{equation} \label{sec_equ_5}
    t_{\alpha}^{\sigma_i} = m \times t_{a}^{\sigma_i}
\end{equation}
\end{enumerate}

\subsection{Simplification of the Event Log}

After simplifying the structures, various networks are generated along with corresponding simplified protocols. Algorithm \ref{Alg3} illustrates this process.
\begin{figure}[htbp]
    \centering
    \includegraphics[width=0.7\textwidth]{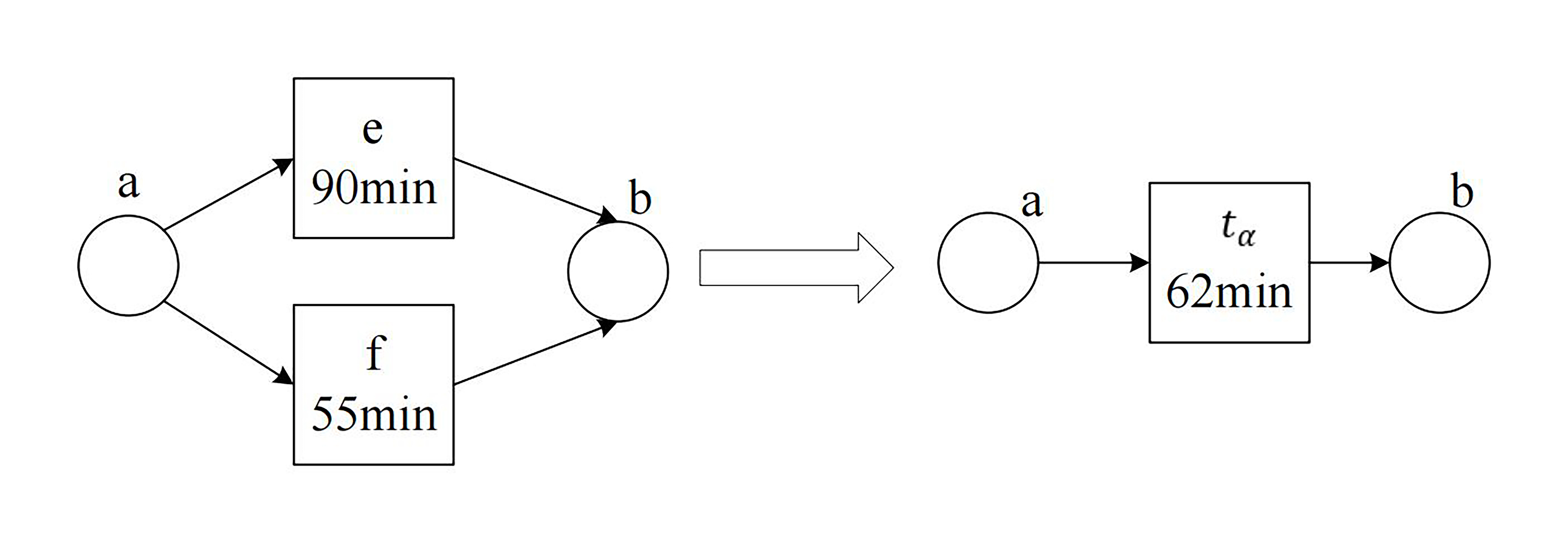} 
    \caption{Or structural sketch.}
    \label{fig:figure 7}
\end{figure}
\begin{algorithm}
    \caption{Simplifying Event Log}\label{Alg3}
        \KwIn{Initial log  $Elog$}
        \KwOut{The simplified event log  $NElog$}
        Get a GSPN structure $N$ from $Elog$\;
        \While {Certain substructures within $N$ can be simplified}{
            \If{\emph{The substructures can be described as a} sequence.}{Using Equation \ref{sec_equ3}, we can create a    new activity denoted as \( t_{\alpha} \)\;}
            \ElseIf {\emph{The substructures can be described as a} or}{Using Equation \ref{sec_equ_4}, we can create a new activity denoted as \( t_{\alpha} \)\;} 
            \ElseIf { \emph{The substructures can be described as a} self-loop}{Using Equation \ref{sec_equ_5}, we can create a new activity denoted as \( t_{\alpha} \)\;} 
            Use \( t_{\alpha} \) to update \( N \) and obtain the new value \( N' \)\;
        }
        Generate the log, denoted as \( NE \log \), of the simplified mesh \( N' \) \;
\end{algorithm}

Figure \ref{fig:figure 7} shows an or substructure. The activities $e$ and $f$ are replaced by a new transition $t_{\alpha}$ for simplicity. The proportion of logs containing $e$ and $f$ in the original log is 20\% and 80\%, respectively. The time delay of $e$ is 90 minutes, and the time delay of $f$ is 55 minutes, then the time delay of the new transition $t_{\alpha}$ is $ 90*0.2+55*0.8=62$.

\subsection{Log Optimization}

Not all simplifications of substructures lead to better predictions. While some simplifications can reduce the event log, they may also significantly impair prediction performance. In this subsection, we aim to optimize the simplification process by focusing on reducing the log size while ensuring reliable prediction performance. First, we evaluate the substructures that need simplification; if we find that a particular substructure will greatly influence future predictions, we choose not to eliminate it. Our optimization approach is guided by the following log optimization formula:

\begin{equation}\label{sec_equ_6}
\begin{split}
    \text{Maximize} \quad & \sum_{i=1}^{n} k_i \times x_i \\
    \text{Subject to:} \quad & \sum_{i=1}^{n} \mu_i \times x_i \leq g \times \Gamma
\end{split}
\end{equation}

In this formulation, \(i\) represents the index of the reducible substructure, taking values from 1 to \(n\). The variable \(x_i\) is a binary constant: it is equal to 1 if the substructure \(N_i\) is reduced, and 0 otherwise. The parameter \(k_i\) corresponds to the number of activities within \(N_i\), while \(\mu_i\) measures the deviation in the remaining time prediction before and after the reduction of \(N_i\). The variable \(\Gamma\) denotes the expected deviation between the predicted and actual values, and \(g\) is a slack variable, typically having a value less than \(n\). 

See Algorithm \ref{Alg4} for more information. The solution to Eq. (\ref{sec_equ_6}) aims to maintain the performance deviation between the final simplified predicted value and the original log within an acceptable range. Refer to Algorithm \ref{Alg4} for additional details.

\begin{algorithm}
    \caption{Optimization of the Log}\label{Alg4}
        \KwIn{$Elog$ and $N^{*} = \{N_1,N_2,\ldots,N_n \}$, $N^{*}$ is the set of reducible substructures. }
        \KwOut {Optimized logs $NElog$}
            To calculate the \( Pr \), please use Algorithm \ref{Alg3}\;
            \For{$N_i$ in $N^*$}{
                    Filter the logs \( Elog \) to extract those related to \( N_i \), resulting in the subset \( Elog_i \)\;
                    Select a prediction point $pr\text{, } pr \in Pr$\;
                    Evaluate the time bias for net structures $N_i$\;
                   \If {by the constraints in Equation (\ref{sec_equ_6})}{update $Elog$ to the simplified new log $NElog$\;}

        }
\end{algorithm}

\section{Experimental Results}\label{sec5}
\subsection{Data Preparation and Forecasting}
The dataset for this study was obtained from the 4TU Centre for Research Data (Mannhardt \& Blinde, 2017) and includes a real-life event log from a Dutch hospital, specifically focusing on cases of the life-threatening disease sepsis. Each entry in the dataset represents a patient’s treatment journey, recorded from their admission to the emergency department until their discharge. The hospital’s ERP system captures the events, which comprise approximately 1,000 instances with a total of 15,000 recorded events. These events document 16 different activities and 39 data attributes, with certain infrequent events (occurring fewer than 10 times) categorized as ``other’’ in this study. Notably, any return visits to the hospital after discharge are excluded from the dataset. The event and attribute values have been anonymized, which includes the omission of details such as the group responsible for each activity, test results, and checklist information. Although the timestamps of the events have been randomized, the interval between events has remained unchanged. This protocol was specifically selected for the simplification experiments described in this paper.

\begin{figure}[htbp]
    \centering
    \includegraphics[]{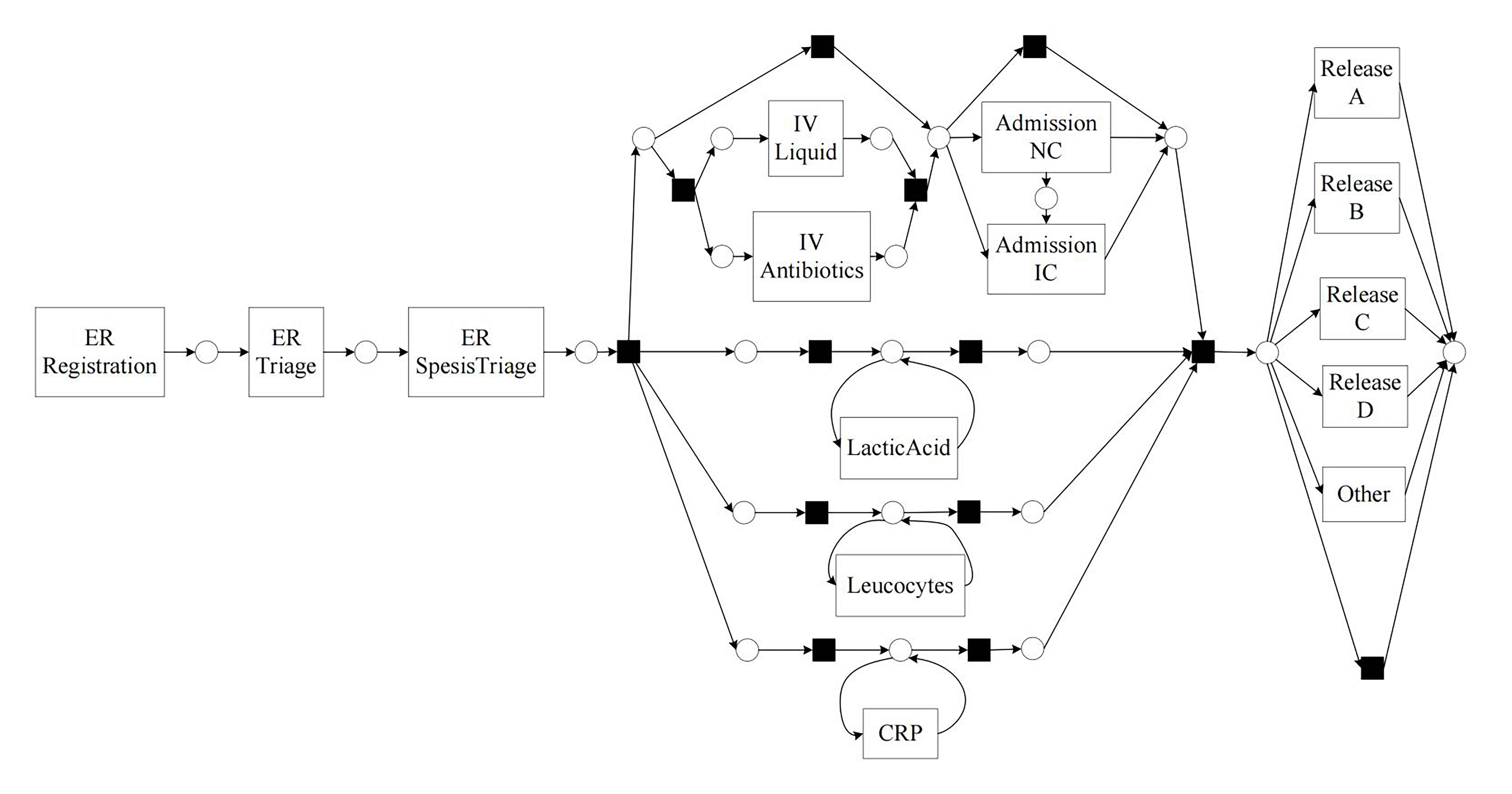} 
    \caption{A GSPN network model for the sepsis event log.}
    \label{fig:figure 8}
\end{figure}

The event log for life-threatening sepsis cases from a Dutch hospital is detailed in Mannhardt and Blinde (2017). An artificially generated ideal model successfully aligns with 98.3\% of the event log, accurately reproducing almost all events. To enhance the validation of the experiments presented in this paper, the model was slightly modified based on the actual situation, resulting in the model depicted in Figure \ref{fig:figure 8}. The event activities included are as follows:
\begin{enumerate}
    \item Three activities related to emergency department registration and triage:
    \begin{itemize}
        \item ER registration
        \item ER triage
        \item ER sepsis triage
    \end{itemize}
    \item Three sepsis-related tests: 
    \begin{itemize}
        \item Leukocytes 
        \item CRP
        \item Lactic acid
    \end{itemize}
    \item Two activities related to admission or transfer to regular care or the ICU: 
    \begin{itemize}
        \item Admission NC
        \item Admission IC 
    \end{itemize}
    \item Five activities related to discharge:
    \begin{itemize}
        \item Release D
        \item Release E (with specific discharge methods labeled as anonymous).
    \end{itemize}
\end{enumerate}
To simulate the use of historical data for training predictive models and applying them to real-world scenarios, this article employs time segmentation to divide event logs into training and test sets. All instances in the log are sorted by their start time. The first 80\% of the sorted logs are designated as the training set, which is used to fit the model, while the remaining 20\% serve as the test set to evaluate predictive accuracy. The predictor or classifier is trained on all instances that begin before a specified date $T_1$ (representing the current time in real-life situations) and only tests instances that start after that date.

In this experiment, we utilized a cluster bucketing method that incorporates a k-means clustering algorithm along with index coding. This approach enables us to predict data after it has been encoded. For our predictions, we employed the Extreme Gradient Boosting (XGB) algorithm. In this study, we conducted basic feature extraction, calculating the execution time of each event in the log by subtracting the timestamp of the previous event from that of the current event. The accuracy of our predictions regarding the remaining time is evaluated by measuring the Mean Absolute Error (MAE) between the predicted and actual results. This paper employs a specific method to quantify prediction errors, accurately reflecting the nature of these errors in predicting remaining time.

\begin{equation}\label{sec_equ_7}
    MAE = \frac{1}{N} \sum_{i=1}^{n} |e_i-\hat{e_i}|
\end{equation}
where $e_i$ is the actual value at the given prediction point, and $\hat{e_i}$ is the predicted value.
\subsection{Application and Analysis}
\begin{figure*}[!htbp]
    \centering
    \includegraphics[]{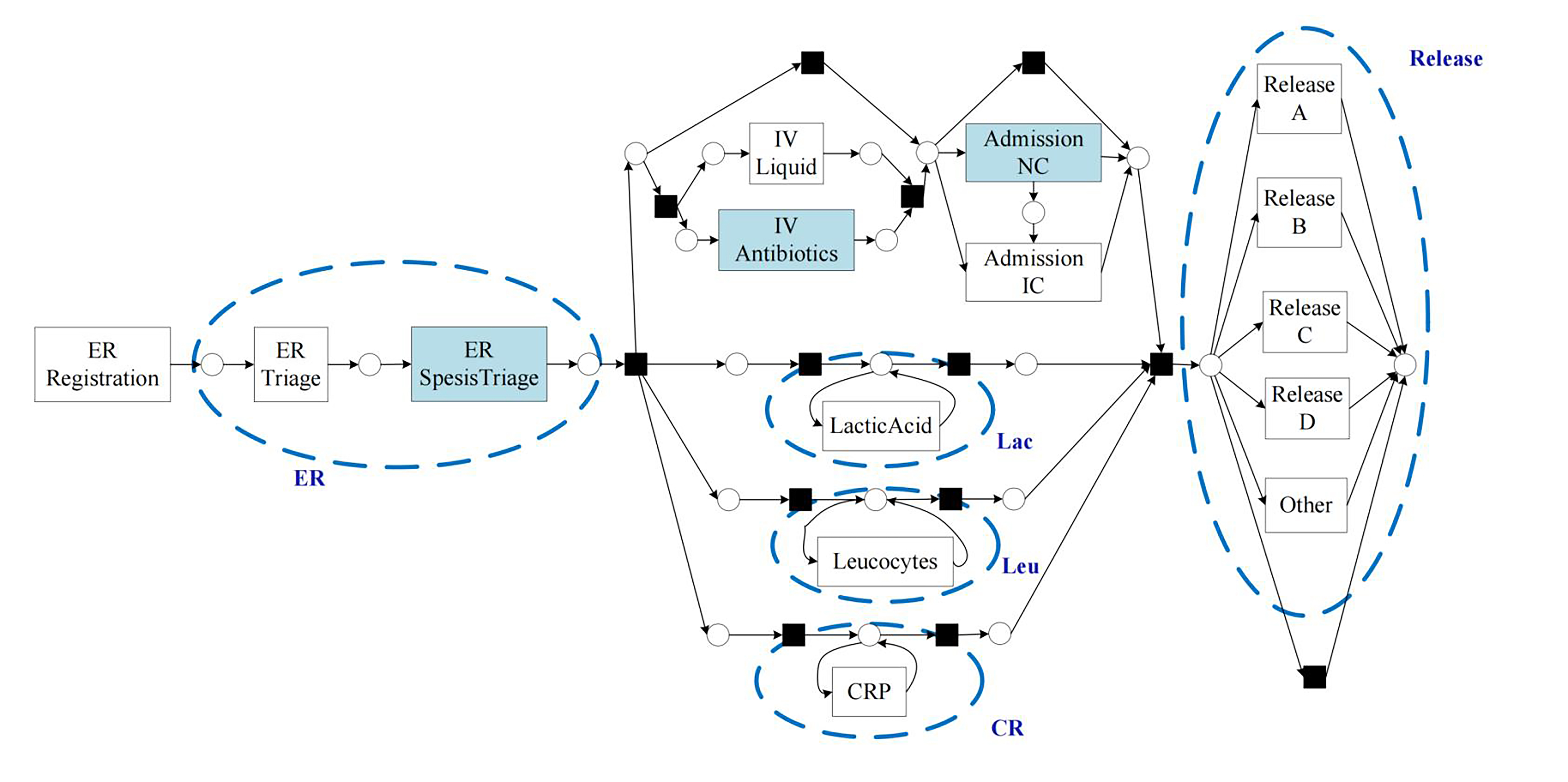} 
    \caption{GSPN remaining time prediction model with reducible structure and prediction points.}
    \label{fig:figure 9}
\end{figure*}

\textbf{1. Simplified substructures and event logs} 
Using Algorithm \ref{Alg2} to select prediction points, the chosen points are ER Sepsis Triage, IV Antibiotics, and Admission NC. This paper identifies several substructures, including sequential structures and structures with self-loops. The subnetwork structures derived from the association matrix method in the event log model for sepsis cases are represented as blue circles in Figure \ref{fig:figure 9}, denoting ER, CR, Lac, Leu, and Release, respectively.

\indent The blue rectangles indicate the prediction points. By applying Algorithm \ref{Alg3} to simplify the event log, we obtain simplified event logs corresponding to the identified substructures. For each prediction point identified, we use 80\% of the original event log's training set as input and implement Algorithm \ref{Alg4} to calculate the correlation between the predicted values and the actual values from the original event log at various prediction points. The results are presented in Table \ref{tab:table 2}. Similarly, the final simplified data for ER, CR, Lac and Leu can be obtained.
\begin{table}
    \centering
    \caption{Difference between predicted and actual values in event logs ($MAE$).}
    \footnotesize
    \setlength{\tabcolsep}{4pt} 
    \renewcommand{\arraystretch}{1.2} 
    \begin{tabular*}{\linewidth}{l|p{2cm}|c|c|c|c|c|p{3cm}}
    \textbf{Prediction point} &\textbf{Original event log} & \textbf{ER(2)} & \textbf{CR(7)} & \textbf{Lac(1.26)} & \textbf{Leu(5.88)} & \textbf{Release(1)} & \textbf{Final reducible part} \\
    \hline
    ER Sepsis Triage & 316613.6 & - & 294207.4 & 343801.4 & 292179.1 & 316924.2 & CR+Lac+Leu+release \\
    IV Antibiotics & 582101.8 & 562752.3 & 734838 & 764518.5 & 473945.4 & 581625.3 & ER+CR+Leu+Lac \\
    Admission NC & 592957.4 & 571664.1 & 662848.9 & 748153.1 & 432429.1 & 588133.7 & ER+CR+Leu+Lac
    \end{tabular*}
    \label{tab:table 2}
\end{table}

\begin{table*}[!]
    \centering
    \caption{Deviation of predicted values and data volume for the final simplified protocol.}
    \footnotesize
    \setlength{\tabcolsep}{4pt} 
    \renewcommand{\arraystretch}{1.2} 
\begin{tabular*}{\linewidth}{l|l|l|p{2cm}|p{2cm}|p{2cm}|l}
    \textbf{Prediction point} &\textbf{Original data} & \textbf{Simplify data} & \textbf{Prediction performance  improvement} & \textbf{Original data volume} & \textbf{Simplify data volume} & \textbf{Data reduction}  \\
    \hline
    ER Sepsis Triage & 454586.2 & 439542.79 & 3.31\% & 13121 & 7962 & 39\%  \\
    IV Antibiotics & 711153.7 & 354328.84 & 50.18\% & 13121 & 6398 & 51\% \\
    Admission NC & 736749 & 389073.55 & 47.19\% & 13121 & 6398 & 51\% 
\end{tabular*}
    
    \label{tab:table 3}
\end{table*}

\begin{table*}
    \centering
     \caption{ Deviation of predicted values from the true values in different prediction points for each event log ($MAE$).}
    \footnotesize
    \setlength{\tabcolsep}{4pt} 
    \renewcommand{\arraystretch}{1.2} 
    \begin{tabular*}{\linewidth}{l|l|l|l|l|l|l|l}
    \textbf{Prediction point} &\textbf{Original event log} & \textbf{ER(2)} & \textbf{CR(7)} & \textbf{Lac(1.26)} & \textbf{Leu(5.88)} & \textbf{Release(1)} & \textbf{Final reducible part} \\
    \hline
    ER Sepsis Triage & 316613.6 & - & 294207.4 & 343801.4 & 292179.1 & 316924.2 & CR+Lac+Leu+release \\
    IV Antibiotics & 582101.8 & 562752.3 & 734838 & 764518.5 & 473945.4 & 581625.3 & ER+CR+Leu+Lac \\
    Admission NC & 592957.4 & 571664.1 & 662848.9 & 748153.1 & 432429.1 & 588133.7 & ER+CR+Leu+Lac
    \end{tabular*}
   
    \label{tab:table 4}
\end{table*}

\begin{table*}[]
    \centering
    \caption{Deviation of predicted values and data volume for the final simplified protocol.}
      \footnotesize
    \setlength{\tabcolsep}{4pt} 
    \renewcommand{\arraystretch}{1.2} 
\begin{tabular*}{\linewidth}{l|l|l|p{3cm}|p{2cm}|p{2cm}|l}
    \textbf{Prediction point} &\textbf{Original data} & \textbf{Simplify data} & \textbf{Prediction performance  improvement} & \textbf{Original data volume} & \textbf{Simplify data volume} & \textbf{Data reduction}  \\
    \hline
    CRP & 758042.09 & 569963 & 24.81\% & 13121 & 9410 & 28\%  \\
    LacticAcid & 688196.15 & 388266.89 & 43.58\% & 13121 & 7772 & 51\% \\
    Leucocytes & 712511.1 & 688923.32 & 3.31\% & 13121 & 9517 & 27\% \\
    Release A & 648953.5 & 121866.93 & 81.22\% & 13121 & 6398 & 51\% \\
    Release C & 223950.45 & 191093.51 & 14.67\% & 13121 & 6398 & 51\% \\
\end{tabular*}
    
    \label{tab:table 5}
\end{table*}

\begin{figure}[!htbp]
    \centering
    \includegraphics[width=0.5\textwidth]{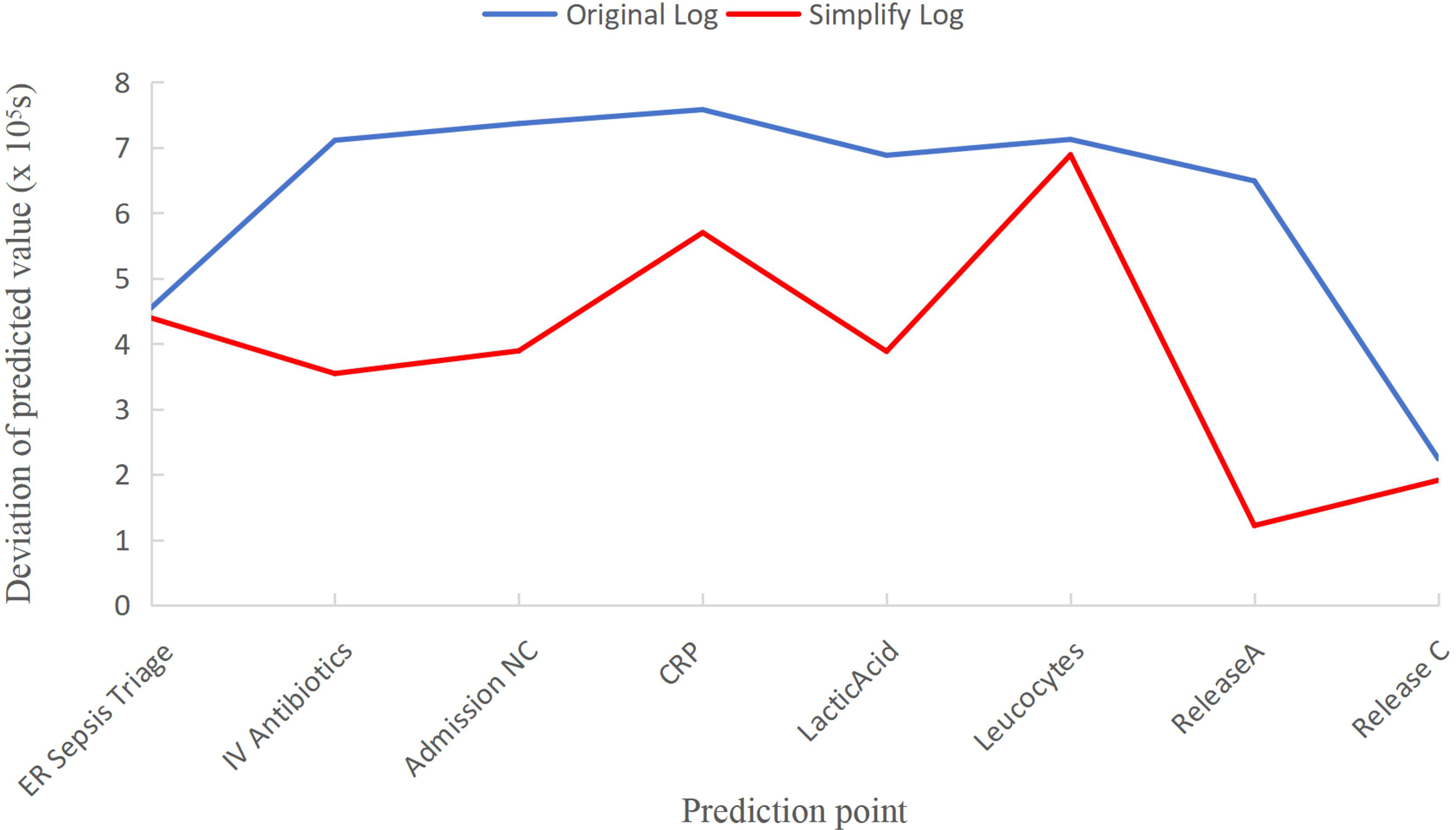} 
    \caption{Deviation from the predicted values for the final simplified protocol. }
    \label{fig:figure 10}
\end{figure}

The comparison between the original event log data and the simplified event log data after reduction was conducted at each prediction point using the test set.
Figure \ref{fig:figure 10} visually represents the differences in prediction performance between the simplified event log and the original event log. Table \ref{tab:table 3} summarizes the deviations between the predicted values and the actual values of the raw data at various prediction points. The final reduced dataset demonstrated improved prediction performance, indicating that the overfitting often observed in the original data due to redundant information was alleviated in the simplified dataset. The variations in prediction performance improvement shown in Table \ref{tab:table 3} are linked to the location of the prediction points within the process model. For instance, if the prediction point is ``ER Sepsis Triage’’, which occurs at the beginning of the process, there is limited prefix data available for analysis.

Data reduction at earlier stages limits the potential for significant improvements in prediction performance. In contrast, the ``Admission NC’’ point occurs later in the process, which enables the collection of more prefix event log data. Consequently, implementing data reduction at earlier stages significantly enhances the predictive performance for ``Admission NC’’.

\begin{figure}[!htbp]
    \centering
    \includegraphics[width=0.5\textwidth]{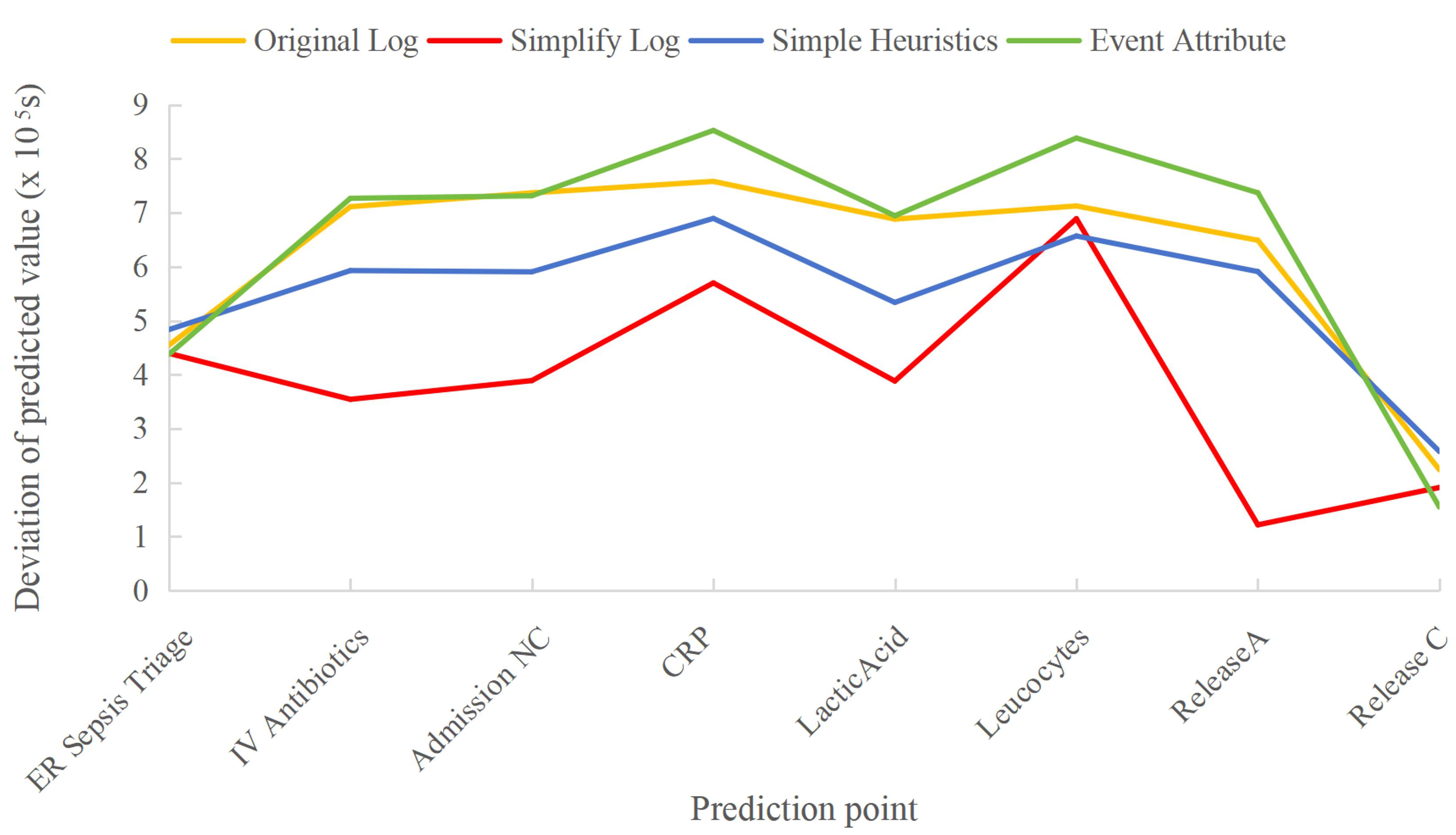} 
    \caption{A comparison of different methods for simplifying event logs and their impact on predictive performance.}
    \label{fig:figure 11}
\end{figure}

To enhance experimental verification, this paper selected CR, Lactic Acid, Leucocytes, Release A, and Release C as additional prediction points for the reduction and prediction experiments. The final results of the reduction are presented in Figure \ref{fig:figure 11}, as well as in Tables \ref{tab:table 4} and \ref{tab:table 5}.

\textbf{2. Comparative study}

Traditional dimensionality reduction methods primarily focus on numerical data and aim to reduce the number of attribute columns. In this paper, we did not directly compare our reduction method with these traditional techniques; instead, we focused on reducing log records. We evaluated the simplified data produced by applying two filters:Filter Log on Event Attribute Values and Filter Log using Simple Heuristics. The first filter targets instances where the CRP value is normal, while the second filter excludes instances that do not start with ``ER Registration’’, ``ER Sepsis Triage’’, or ``ER Triage’’, and those that do not end with ``Release A’’, ``Release B’’, ``Release C’’, or ``Release D’’. The experimental results are illustrated in Figure \ref{fig:figure 11}, which shows the deviation values between the predicted residual times of the reduced logs generated by our proposed method, the simplified logs obtained from the two filters, and the actual values from the original event log used in the experiment. The red line in the figure represents the results from our proposed method. Our findings indicate that the proposed reduction method more effectively preserves the predictive quality of the remaining time than merely deleting data based on specific features.

\section{Conclusion}\label{sec6}

This paper presents a method for simplifying event logs to improve the accuracy of remaining time predictions. Recognizing the significance of resources in estimating various performance metrics, our approach selects prediction points based on resource community networks. This allows us to avoid oversimplifying critical work points. We then apply structured reduction rules to create reduced logs for each substructure, followed by using a remaining-time prediction algorithm to predict and optimize failure values. Our approach results in simplified event logs that can maintain or even enhance the accuracy of remaining time predictions. The experiments confirm that the proposed method achieves optimal simplification of event logs while preserving the accuracy of our remaining time predictions. Although this paper primarily presents the log reduction method in the context of remaining time prediction, our ongoing work aims to demonstrate the effectiveness of the proposed event log reduction framework for predicting performance from various perspectives, such as prediction outcomes or risks. In addition, more research is required to improve the accuracy of the prediction through a more thorough data analysis.

\section*{Acknowledgement}
The authors thank the Fujian Provincial Department of Science and Technology, China, for funding this work through the Science and Technology Planning Project under Grant 2024H0014 (2024H01010100).
\bibliographystyle{fundam}
\bibliography{bibliography}
\end{document}